\documentclass[10pt,twocolumn,letterpaper]{article}

\pdfoutput=1
\usepackage{iccv}
\usepackage{times}
\usepackage{epsfig}
\usepackage{graphicx}
\usepackage{amsmath}
\usepackage{amssymb}
\usepackage{authblk}
\usepackage{multirow}
\usepackage{adjustbox}
\usepackage{tabularray}
\usepackage{placeins}
\usepackage[normalem]{ulem}
\usepackage{float}
\usepackage{booktabs}
\usepackage[dvipsnames,table]{xcolor}

\usepackage[breaklinks=true,bookmarks=false]{hyperref}

\iccvfinalcopy 

\def\iccvPaperID{****} 

\begin{document}

\title{Towards Unbiased Cross-Modal Representation Learning for \\ Food Image-to-Recipe Retrieval}

\author{Qing Wang}
\author{Chong-Wah Ngo}
\author{Ee-Peng Lim}

\affil {Singapore Management University}




\maketitle

\begin{abstract}
This paper addresses the challenges of learning representations for recipes and food images in the cross-modal retrieval problem. As the relationship between a recipe and its cooked dish is cause-and-effect, treating a recipe as a text source describing the visual appearance of a dish for learning representation, as the existing approaches, will create bias misleading image-and-recipe similarity judgment. Specifically, a food image may not equally capture every detail in a recipe, due to factors such as the cooking process, dish presentation, and image-capturing conditions. The current representation learning tends to capture dominant visual-text alignment while overlooking subtle variations that determine retrieval relevance. In this paper, we model such bias in cross-modal representation learning using causal theory. The causal view of this problem suggests ingredients as one of the confounder sources and a simple backdoor adjustment can alleviate the bias. By causal intervention, we reformulate the conventional model for food-to-recipe retrieval with an additional term to remove the potential bias in similarity judgment. Based on this theory-informed formulation, we empirically prove the oracle performance of retrieval on the Recipe1M dataset to be MedR=1 across the testing data sizes of 1K, 10K, and even 50K. We also propose a plug-and-play neural module, which is essentially a multi-label ingredient classifier for debiasing. New state-of-the-art search performances are reported on the Recipe1M dataset.
\end{abstract}
\section{Introduction}
Cross-modal recipe retrieval is potentially a more scalable approach than training a classifier for food image or ingredient recognition~\cite{min2019survey, zhu2021learning}. Given a food image query, the task is to retrieve the corresponding recipe to provide the necessary information, such as dish name and ingredients, for the estimation of nutrition and calorie content. In the literature~\cite{salvador2017learning, carvalho2018cross, chen2018deep, fu2020mcen, zhu2019r2gan, wang2019learning, zhu2020cookgan, sugiyama2021cross, guerrero2021cross, salvador2021revamping, shukor2022transformer}, the above task is posed as a cross-modal representation learning problem, where every recipe and its food image are transformed and projected to a shared embedding space to maximize their pairwise similarity. Under this formulation, a recipe is assumed to provide textual descriptions of the food's visual content. The relationship between a recipe and its food image, however, is beyond visual-textual alignment: a recipe narrates the process of preparing a dish while a food image visualizes the outcome of cooking. The cause-and-effect alignment between a textually described cooking process and a static visual snapshot of the cooking outcome, in general, characterizes the underlying challenge of recipe retrieval using an image as the query. 

We postulate that, due to pairwise alignment, the joint space tends to capture the dominant content of food images and recipes, while overlooking the subtle visual changes due to the cooking process. For example, the recipe ``drunken chicken" lists the {major} ingredient ``chicken" and other {minor} ingredients such as ``ginger" and ``goji". While the major ingredient will visually present in the food image, the minor ingredients such as ``ginger" or ``goji" may not be visible subject to their sizes, occlusion, and image-capturing conditions. Due to the inconsistency between a recipe and its paired image, learning image representations that can capture all details of ingredients for alignment is practically difficult. Therefore, the existing representation learning techniques fall short in capturing the subtle variations to distinguish the recipes of visually similar food images (e.g., drunken chicken, steamed ginger chicken, herbal chicken). 

This paper addresses the bias in representation learning from a causal view that treats food image as the {\em effect} of manipulating culinary elements by following cooking instructions in a recipe. The causal view allows us to identify the potential confounders that introduce spurious correlation in representation learning. In general, culinary elements such as ingredients, cooking, and cutting methods are potential confounders. In this paper, we mainly consider ingredients to formulate the problem. By backdoor adjustment~\cite{pearl2009causality}, a causal-informed equation is derived to alleviate the potential bias arising from ingredients. We discuss the challenge of realizing this equation for recipe retrieval and propose a neural module implementation to approximate this equation. The module is essentially a multi-labeled ingredient classifier that predicts the distribution of ingredients in an image for readjustment of image representation during learning. This debiasing module can be readily plugged into most of the existing neural architectures for recipe retrieval, including the SoTA models such as H-T~\cite{salvador2021revamping}, TFood~\cite{shukor2022transformer}, VLPCook~\cite{shukor2022structured} for cross-modal learning.

A significant contribution of our work is that the causal view provides a causal theory-informed upper-bound performance for the problem of image-to-recipe retrieval. By simulating the proposed debiasing module with the desired classification performance, a near-perfect retrieval performance is attained on the Recipe1M dataset~\cite{salvador2017learning}. When applying the module to debiase the representation learning of SoTA models such as VLPCook~\cite{shukor2022structured}, the new state-of-the-art performances are also reported on Recipe1M. To the best of our knowledge, this is the first paper addressing the bias in representation learning based on the causal theory for cross-modal recipe retrieval.
\section{Related Work}
The past research efforts in recipe retrieval are mostly devoted to learning to represent the highly structured information in recipes~\cite{chen2017cross1,salvador2017learning}. The early works~\cite{carvalho2018cross,salvador2017learning} have established the basis for encoding the three sections of information in a recipe (dish title, ingredients, instructions) independently using LSTM. The encoded representations of each section are then concatenated and projected to align with the visual representation. Variants of methods have been proposed based on this LSTM-based architecture, including attentional mechanism~\cite{chen2018deep, wang2021cross, li2021multi}, multi-task learning~\cite{chen2017cross2}, hierarchical modeling of semantics~\cite{pham2021chef}, and adversarial learning~\cite{zhu2019r2gan, wang2019learning, guerrero2021cross, zan2020sentence}. As reported by a comparative study in~\cite{zhu2021learning}, these efforts, which explore textually rich descriptions in recipes to align food images, have incrementally boosted search performance. 

Transformer-based neural architectures have also been investigated for image-and-recipe alignment. Representative works include H-T (hierarchical recipe transformer)~\cite{salvador2021revamping}, T-Food (transformer decoders for food)~\cite{shukor2022transformer} and VLPCook (visual-language pre-training)~\cite{shukor2022structured}. This line of works is first demonstrated by H-T~\cite{salvador2021revamping} via replacement of LSTM with a transformer to demonstrate significant improvement in search. Following early works, H-T also encodes the three sections of the recipe independently. T-Food~\cite{shukor2022transformer} improves H-T by having additional layers of transformers to model interactions among the representations encoded from the title, ingredient, and cooking instruction sections. Furthermore, different from~\cite{salvador2017learning,carvalho2018cross, fu2020mcen, xie2021learning} which typically employ ResNet-50 as an image encoder, T-Food employs transformers to encode image features as well as modeling the inter-dependencies between image and recipe features. More recently, foundational models have also been employed in~\cite{shukor2022structured,huang2023improving}. VLPCook~\cite{shukor2022structured} leverages CLIP~\cite{radford2021learning} to extract the local (ingredient) and global (cooking instructions) contexts from food images. Vision transformer (ViT) is then employed to jointly embed the food images and contexts for image representation learning. 

Despite these efforts to align the complex cooking process with food images, the joint space is established based on the objective of maximizing the similarity (or correlation) between the image-recipe pairs. There are few studies investigating the potential biases in representation learning for recipe retrieval, as in this paper. The most closely related works are~\cite{sugiyama2021cross, kim2021learning}, which studies feature disentanglement to reduce database bias. In~\cite{sugiyama2021cross}, RDE-GAN disentangles recipe representation from dish presentation style, which is irrelevant to the cooking process. The dish style features, which are explicitly extracted from food images during training, are segregated from image representation while learning to align with recipes. In~\cite{kim2021learning}, Retrieval-IVAE employs identifiable-VAE (variational autoencoder) to learn the semantics of latent factors in representation learning. This study shows the potential of learning disentangled representations to explain the variability of paired multimodal data. Different from RDE-GAN~\cite{sugiyama2021cross} and Retrieval-IVAE~\cite{kim2021learning}, our work addresses the learning bias (or spurious correlation) from the causal view that suggests potential confounders. This novel view provides a theory-informed way of removing ambient factors (e.g., dish style, image capturing conditions) mentioned in~\cite{li2021cross, sugiyama2021cross}. It is worth mentioning that there are also efforts aiming to learn robust representation by reconstruction of cooking programs~\cite{papadopoulos2022learning} and recipes~\cite{salvador2019inverse, chhikara2024fire, wang2022learning} from images. Although these approaches are not framed based on the causal theory, the multi-modal representations learnt in such a way might also capture the causal effects of cooking. 

Causal inference has been applied for representation learning~\cite{wang2022learned, yang2021deconfounded, zhang2020causal}. Nevertheless, different from our work, most of the studies apply causal adjustment for single-modal image classification problems~\cite{chen2022towards, tang2020long, mahajan2021domain}. These works focus on the removal of bias from pretrained models for domain adaptation~\cite{stojanov2021domain}. Different from our work that explores causality-based learning on cross-modal paired data, these works usually leverage plenty of visual examples associated with a class label for debiasing. There are also a few works~\cite{ma2022ei, yang2021deconfounded} that address the bias on learning from paired data. However, these works assume there is a direct correspondence between captions and visual entities for debiasing. Furthermore, as these works aim for the removal of bias in the pretrained models, the confounder is assumed latent or unobserved. As a result, additional heuristics are required for front-door adjustment to remove bias. Our work is not for the removal of the common-sense bias in pretrained models as in~\cite{ma2022ei, yang2021deconfounded}. As confounder can be identified, it leads to an elegant way of simulating oracle search performance.  
\section{Causality-based Representation Learning}
The major ingredients of a dish are often presented to be more visible than the minor ingredients. Furthermore, some ingredients may be occluded and even dissolve during cooking. It becomes unrealistic to assume that a food image can capture the visual property of ingredients recorded in a recipe equally. We argue that the existing representation learning sacrifices {the minor and partially occluded details} to compensate for the learning of a joint space, resulting in a bias in similarity measurement.

\begin{figure}
    \centering
    \includegraphics[width=8cm]{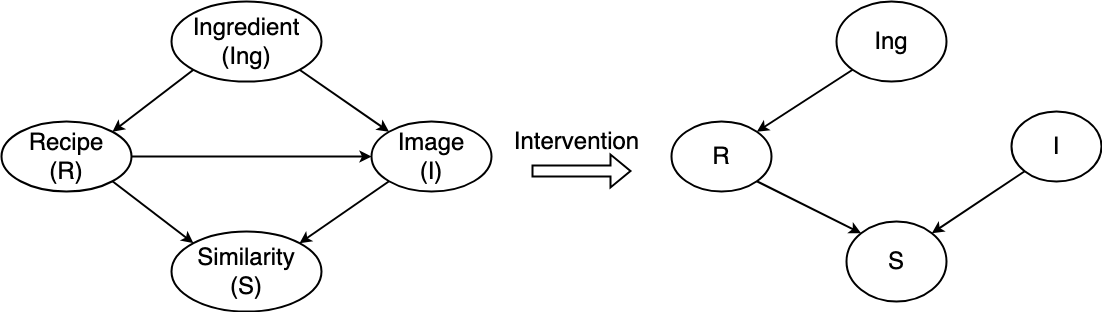}
    \caption{Left: A causal graph depicting how cross-modal similarity is affected by the spurious correlation in learning image and recipe representations due to the confounder $Ing$. Right: Debiasing by backdoor adjustment to cutoff the incoming edges to the image.}
    \label{fig:causal_graph}
\end{figure}

\subsection{A Causal View}
Ingredients are the main component of a recipe, affecting the composition of a dish (and the content of an image). We imagine that a dish image is a product of cooking, by following the instructions of a recipe to process the ingredients. Let $I$, $R$, and $Ing$ denote the image, recipe, and ingredient, respectively. Their triple relationship (i.e., $I$, $R$, $Ing$) can be depicted as a directed graph shown in Figure~\ref{fig:causal_graph}, where $Ing$ is the confounder of $R$ and $I$. The arrow in Figure~\ref{fig:causal_graph} denotes a casual direction. For example, $Ing \to R$ means that a recipe is written based on (or caused by) ingredients; $Ing \rightarrow I \leftarrow R$ means $I$ is produced by (or caused by) both $Ing$ and $R$. The confounder $Ing$ complicates the flow of information through the pathways $R \leftarrow Ing \rightarrow I$ and $Ing \rightarrow R \rightarrow I$, creating spurious correlation during representation learning~\cite{pearl2009causality}. Particularly, the major ingredients that are expected to be more visible in the final cooked image, and less likely to be occluded due to dish presentation or image capturing angle, will have more influence on representation learning. Consequently, the similarity assessment (denoted as $S$ in Figure~\ref{fig:causal_graph}) between $I$ and $R$ is dominated by the major or popular ingredients in a dish. Failure to capture the non-major ingredients that describe the subtle variations in a dish is often a cause of imprecise cross-modal similarity measurement~\cite{zhu2021learning}. 

By Bayes rule, the similarity, represented as the conditional probability $P(S|I, R)$, can be described as:
\begin{subequations} \label{eq:biasing_equation}
    \begin{align}
        P(S|I,R) & = \sum_{ing}P(S,ing|I,R) \label{eq:existng_1} \\
       & = \sum_{ing}P(S|I,R,ing){P(ing|I,R)} \label{eq:existng_2},
    \end{align}
\end{subequations}
where the term, $P(ing|I,R)$, in Eq.~(\ref{eq:existng_2}) indicates that the probability of an ingredient, which is conditioned on the image-recipe pair, will impact the similarity assessment. An intuitive explanation of why this term could cause spurious correlation is as follows. In general, the appearance of a dish, which is prepared based on a recipe, is the effect of applying the cooking steps to the ingredients. However, appearance varies across persons, cultures, and environments. For example, a dish cooked by different persons may be visually different due to personal style in cooking, dish presentation, and image capturing angles and lighting conditions. As a result, the visibility of ingredients i.e., $P(ing|I, R)$, may differ across the dishes cooked with the same recipe. Furthermore, the same ingredient (e.g., onion) may appear in one dish but invisible in another dish due to different cooking methods. Learning the correspondence between image and recipe representation becomes dataset-dependent.

Our idea to remove the spurious correlation is by transforming the ingredient probability to be induced directly from the recipe or the image, i.e., $P(ing|R)$ or $P(ing|I)$, rather than based on their co-existence in both image and recipe representations. Using $P(ing|R)$ as an example, this transformation is equivalent to backdoor adjustment in causal theory~\cite{pearl2009causality}, by removing all the incoming edges to image $I$, as shown in Figure~\ref{fig:causal_graph}.

\subsection{Backdoor Adjustment}\label{sec:backdoor_adjustment}
The main idea to address the spurious correlation in representation learning is by intervening the image variable, i.e., $do(I)$, such that the information flows from $Ing$ and $R$ are cutoff, as shown in Figure~\ref{fig:causal_graph} (right). The intervened version of the similarity measure is:
\begin{subequations} \label{eq:debiasing_equation}
    \begin{align}
 & P(S|do(I),R) \notag \\
         & = \sum_{ing} P(S|do(I),R, ing)P(ing|do(I),R) \label{eq:debiasing_1} \\
       & = \sum_{ing} P(S|do(I),R, ing)P(ing|R) \label{eq:debiasing_2} \\
       & = \sum_{ing} P(S|I,R,ing){P(ing|R)} \label{eq:debiasing_3},
    \end{align}
\end{subequations}
where Eq.~(\ref{eq:debiasing_1}) is obtained in the same way as Eq.~(\ref{eq:existng_2}) by following the Bayes rule. By the rule-3 of $do$-calculus (Theorem 3.4.1 ~\cite{pearl2009causality}), the $do(I)$ in the term $P(ing|do(I),R)$ can be omitted. This is due to $S$, which is a collider of $R$ and $I$, that blocks the information flow from $Ing$ to $I$, i.e.,  $Ing \rightarrow R \rightarrow S \leftarrow I$. Hence, $P(ing|do(I),R)=P(ing|R)$. Subsequently, by rule-2 of do-calculus, we arrive at Eq.~(\ref{eq:debiasing_3})  because $S$ is independent of $I$ after removing the outgoing edges from $I$. 

\subsection{Neural Implementation}
To this end, Eq.~(\ref{eq:debiasing_3}) has adjusted the bias by weighting the similarity with the true distribution of ingredients in a recipe, i.e., $P(ing|R)$ rather than $P(ing|R, I)$. Next, we describe the implementation of Eq.~(\ref{eq:debiasing_3}) using a neural network. As similarity is often implemented as a dot product operation, we set $P(S|I, R, ing) = f_s(e_I,e_R, e_{ing})$, where $e_I, e_R, e_{ing}$ denote the embedding of $I$, $R$ and $ing$, respectively. We have:
\begin{subequations} \label{eq:implement_equation1}
    \begin{align}
    & P(S|do(I),R) \notag \\
     & = \sum_{ing} f_s(e_I,e_R, e_{ing}){P(ing|R)} \label{eq:implement_eq1} \\
    & = \mathbb{E}_{[ing|R]}[f_s(e_I,e_R, e_{ing})] \label{eq:implement_eq2} \\
    & = \mathbb{E}_{[ing|R]}[e_R \cdot (e_I + e_{ing})] \label{eq:substitute_equation} \\
    & = e_R \cdot (e_I + \mathbb{E}_{[ing|R]}[e_{ing}]) \label{eq:implement_eq3}  \\
       & = e_R \cdot (e_I + \sum_{ing}P(ing|R) \cdot e_{ing}) \label{eq:implement_eq4} \\
       & = \underbrace{e_R \cdot e_I}_{{similarity}} + \underbrace {e_R \cdot \sum_{ing}P(ing|R) \cdot e_{ing}}_{debiasing} \label{eq:implement_eq5},
    \end{align}
\end{subequations}
where Eq.~(\ref{eq:implement_eq2}) is expressed based on the definition of expectation, and the similarity function is implemented as $f_s(e_I,e_R, e_{ing})=e_R \cdot (e_I + e_{ing})$ in Eq.~(\ref{eq:substitute_equation}). Note that a similar implementation is also adopted by~\cite{qi2020two}. Thanks to the linear property of expectation, the equation can be simplified to Eq.~(\ref{eq:implement_eq3}), by moving the expectation inside the parenthesis. Further expanding the equation will arrive at Eq.~(\ref{eq:implement_eq4}) and Eq.~(\ref{eq:implement_eq5}). 

Eq.~(\ref{eq:implement_eq5}) expands the conventional similarity measure (i.e., the dot product term)~\cite{salvador2021revamping, shukor2022transformer} with an additional debiasing term. There are two challenges in implementing Eq.~(\ref{eq:implement_eq5}). First, the additional term requires enumeration of $P(ing|R)$ over all the ingredients being considered. For a dataset with tens of thousands of ingredients (e.g., Recipe1M~\cite{salvador2017learning}), the overhead is considerably high. We engineer the problem by calculating the expectation in debiasing term over a subset of ingredients, which will be further detailed in Section~\ref{sec:framework}. 

Second, the backdoor adjustment in Eq.~(\ref{eq:debiasing_3}) is established on the fact that $R$ is the ground-truth recipe of $I$. Specifically, in Eq.~(\ref{eq:implement_eq5}), $P(ing|R)$ cannot be computed directly as $R$ is the search target and is unknown during query time. Fortunately, as both image and recipe representations are learned to be similar of each other, it is still feasible to predict the ingredients of $R$ from $I$. In other words, we set $P(ing|R) \approx P(ing|I)$ in Eq.~(\ref{eq:implement_eq5}) as following:
\begin{subequations} \label{eq:implement_equation1}
    \begin{align}
    & P(S|do(I),R) \notag \\
       & \approx e_R \cdot e_I + e_R \cdot \sum_{ing}P(ing|I) \cdot e_{ing} \label{eq:approx_eq1} \\
       & = e_R \cdot \underbrace{(e_I + \mathbb{E}_{[ing|I]}[e_{ing}] )}_{image\ debiasing} \label{eq:approx_eq2} \\
       & = e_R \cdot \widetilde{e}_I \label{eq:approx_eq3}.
    \end{align}
\end{subequations} 
In Eq.~(\ref{eq:approx_eq2}), the image representation is debiased with $\mathbb{E}_{[ing|I]}[e_{ing}]$, which performs adjustment by a linear sum of ingredient embeddings weighted by their probabilities. The proxy of $R$, i.e., $P(ing|I)$, decodes the ingredients observed from $I$ to augment $e_I$. As a consequence, $\widetilde{e}_I$ is explicitly enhanced with missing details, e.g., minor and partially occluded ingredients, overlooked during representation learning. The representation $\widetilde{e}_I$ is also resilient to non-popular ingredients that might not be captured properly during learning. In our experiment (Section~\ref{sec:exp_robust}), we notice that $\widetilde{e}_I$ is also robust in retrieving the recipes of food categories unseen during training.
\begin{figure*}[t]
\centering
\includegraphics[scale=0.3]{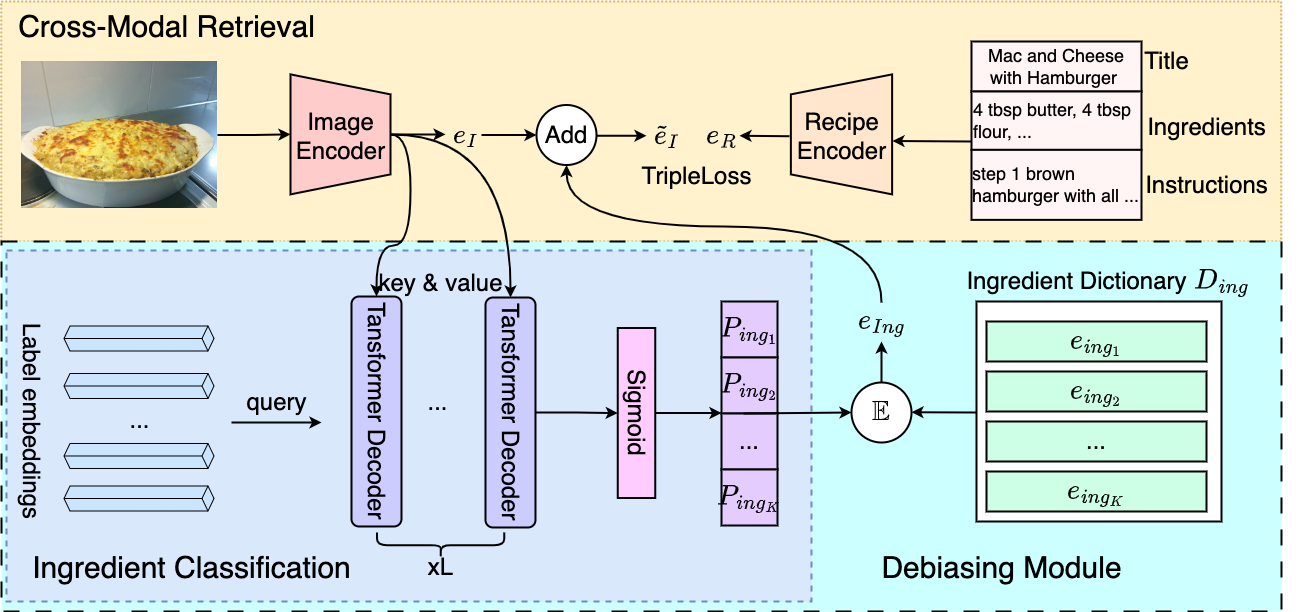}
\caption{There are three components in our framework: image embedding $e_I$ generated by the image encoder, recipe embedding $e_R$ generated by the recipe encoder, and debised image embedding $\widetilde{e}_I$. The triplet loss $L_{triplet}$ is applied on $\widetilde{e}_I$ and $e_R$. The proposed retrieval debiasing module is illustrated on the right. We utilize the Transformer decoder for our ingredient classification, which takes image embedding $e_I$ as key and value, and each ingredient label embedding as a query. We apply the sigmoid function to the output embedding from the last layer of the Transformer decoder and obtain ingredient prediction probabilities $P_{ing}$. By multiplying the probabilities with the ingredients in dictionary $D_{ing}$, we get the expectation of ingredient embedding $\mathbb{E}_{[ing|I]}[D_{ing}]$. The debiasing image embedding $\widetilde{e}_I$ is obtained by adding $e_I$ and $\mathbb{E}_{[ing|I]}[D_{ing}]$. We train the Transformer decoder using asymmetric loss~\cite{Ridnik_2021_ICCV}.}
\label{fig:framework}
\vspace{-0.3cm}
\end{figure*}

\section{Cross-Modal Recipe Retrieval}\label{sec:framework}
The overall architecture of the retrieval framework is depicted in Figure~\ref{fig:framework}. Similar to other existing approaches~\cite{salvador2021revamping, shukor2022transformer, shukor2022structured}, the framework is composed of visual and textual encoders to separately embed the image and recipe into a joint space. The key difference is the additional debiasing component, which implements $P(ing|I)$ using a multi-label classifier and then retrieves relevant ingredients from an ingredient dictionary. Note that the debiasing component can be plugged into most of the existing frameworks for image-to-recipe retrieval~\cite{salvador2021revamping, shukor2022transformer, shukor2022structured}. Here, we use H-T~\cite{salvador2021revamping} as an example to explain the implementation.

\textbf{Encoders} The image encoder is implemented with ResNet-50~\cite{he2016deep} pretrained on ImageNet dataset. The recipe encoder is composed of three transformers to encode the tokenized sentences of dish title, ingredients, and cooking steps in a recipe, respectively. Note that two-level hierarchical transformers are employed to encode ingredient lists and cooking steps, which are treated as sentence sequences. The three embeddings are concatenated and then projected to the joint space as a representation similar to and in the same dimension as the image counterpart.

\textbf{Debiasing} module implements the formula $\mathbb{E}_{[ing|I]}[e_{ing}]$ in Eq.~(\ref{eq:approx_eq2}). A multi-label classifier~\cite{liu2021query2label} is employed to predict the probability distribution of ingredients. Specifically, the classifier is a Transformer decoder, taking $e_I$ as both the key and value while utilizing learnable tokens as queries for cross-attention modeling. With the sigmoid activation function, only ingredients with a probability larger than 0.5 are considered for debiasing. We normalize the probabilities of the chosen ingredients such that their sum is 1. With these, the corresponding ingredient embeddings are sampled from the dictionary and then linearly summed with the ingredient probabilities as their weights, as shown in Figure~\ref{fig:framework}. Note that the debiasing module can be implemented with any ingredient classifiers~\cite{chen2016deep, min2016being, salvador2019inverse}.

\textbf{Ingredient Dictionary} The ingredient dictionary is a static storage containing the embeddings of different ingredients. Recall that the recipe encoder has three transformers, and the ingredient transformer is employed to construct the dictionary. Specifically, the embedding of an ingredient is obtained by averagely pooling the embeddings of an ingredient extracted by the transformer from different recipes. Hence, each entry in the dictionary stores a mean representation of an ingredient. To reduce the size of the dictionary as well as to shorten the time in performing $\mathbb{E}_{[ing|I]}[e_{ing}]$, only the top $K$ most popular ingredients of a dataset are retained, where $K=500$. Note that the dictionary includes seasoning and condiments which are minor ingredients in most recipes.

\textbf{Training objective} The overall training loss includes triple and classification losses for learning joint space embedding and multi-label classifier, respectively. The triple loss ${\mathcal L_{triple}}$, specifically the bi-directional triple loss~\cite{wang2019learning}, is employed to ensure that the pairs of image and recipe embeddings are closely resembled.  The classification loss is based on the asymmetry loss proposed in~\cite{Ridnik_2021_ICCV}, a variant of focal loss, for addressing the adverse effect of learning on long-tailed distribution data such as ingredients. Specifically, given the ingredient prediction probability $p = [ p_{ing_1},  p_{ing_2}, ..., p_{ing_k}]$ of an image $I$, the loss function for an image $I$ is as follows:
\begin{equation}
    \mathcal{L}_{I}=\frac{1}{K} \sum_{k=1}^K \begin{cases}\left(1-p_{ing_k}\right)^{\gamma+} \log \left(p_{ing_k}\right), & y_{ing_k}=1, \\ \left(p_{ing_k}\right)^{\gamma-} \log \left(1-p_{ing_k}\right), & y_{ing_k}=0,\end{cases}
    \label{eq:asymmetric_loss}
\end{equation}
where $y_{ing_k}$ indicates the presence of the ingredient. The parameters $\gamma+$ and $\gamma-$ are to weight positive and negative samples differently in focal loss and are empirically set to $\gamma+ = 1$ and $\gamma-=1$. The overall classification loss $\mathcal{L}_{cls}$ is then calculated by averaging over all the training examples. To this end, the objective function is a weighted linear combination of two losses:
\begin{equation}  \label{eq:total_loss}
    \mathcal{L} = \mathcal{L}_{triple} + \lambda_{cls} \mathcal{L}_{cls},
\end{equation}
where $\lambda_{cls}=0.001$ is a hyperparameter to balance the triple loss and classification loss. Considering that the classification loss will introduce perturbation to representation learning, thus, the $\lambda_{cls}$ is set to a relatively small value, or otherwise the learning of joint space will take a longer time to converge and even becomes suboptimal.

\textbf{Training Procedure} We first train the image and recipe encoders as in~\cite{salvador2021revamping}. The trained ingredient transformer is then utilized to extract ingredient features for building the dictionary. With these, the framework in Figure~\ref{fig:framework} is composed for the next round of end-to-end training.  {Specifically, the weights of encoders and the dictionary are fine-tuned based on the loss functions}, while the multi-label classifier is trained from scratch to debiase representation learning.
\section{Experiments}
The experiments are conducted on Recipe1M dataset~\cite{salvador2017learning}. A total of 238,999 image-recipe pairs are used for model training, together with another set of 51,119 pairs for validation. The search sets are formed by sampling recipes from the remaining 51,303 pairs not in the training and validation sets. For each set, we sample randomly 10 times and report average performance. We sample search sets of different sizes in the multiple of 10K. Unless otherwise stated, we follow the existing evaluation protocol~\cite{salvador2021revamping} to report the performance on 1K and 10K test sets. The evaluation metrics are median rank and Recall@K, where K=\{1,5,10\}. The former, abbreviated as medR, measures the median rank of all search targets, with medR=1 indicating at least half of the targets ranked at the top position. The latter, abbreviated as R1, R5, and R10, evaluates the ability of a model to rank a target within a search depth of K. Note that, for a retrieval model, a lower value of medR and a high value of recall are preferred. 

\textbf{Implementation details.} 
We verify the proposed debiasing module on top of various SoTA models reported in the literature, including H-T~\cite{salvador2021revamping}, TFood~\cite{shukor2022transformer}, VLPCook~\cite{shukor2022structured}. These models use different image encoders: ResNet-50 for H-T, ViT-B/16 for TFood, and CLIP-ViT-B/16~\cite{radford2021learning} for VLPCook. The former two encoders are pretrained on ImageNet, while the latter is initialized with weights of CLIP. For the recipe encoder, all the models use a transformer with 2 layers and 4 heads. We strictly follow the original implementations of these models, except plugging in the debiasing module as the example shown in Figure~\ref{fig:framework}. The classifier in the debiasing model is implemented based on the transformer architecture similar to~\cite{liu2021query2label}, consisting of one encoder layer and two decoder layers with each having 4 heads. All images are resized to the resolution of $256 \times 256$ and then cropped to $224 \times 224$. The model is trained with a batch size of 64 using Adam optimizer with a base learning rate $10^{-4}$ for H-T and $10^{-5}$ for TFood and VLPCook. The dictionary has entries for 500 ingredient representations. Every recipe in Recipe1M has at least one ingredient that can be retrieved from the dictionary.

\subsection{Oracle Performance}
\begin{figure}
    \centering
    \includegraphics[width=\columnwidth]{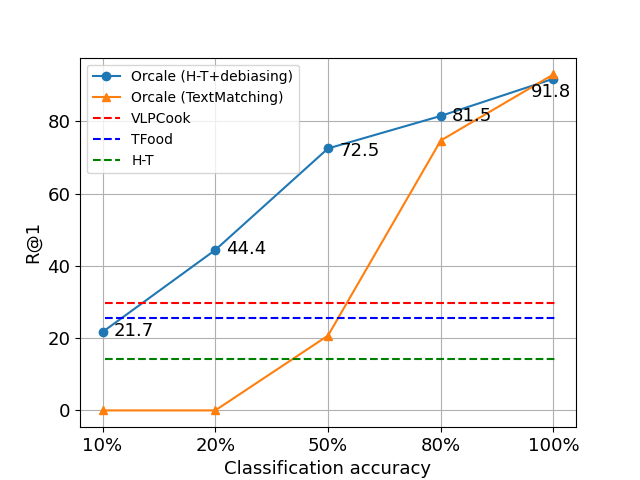}
    \caption{Recall@1 for image-to-recipe retrieval on 50K test set by varying the accuracy of ingredient prediction. The solid lines are oracle runs. The dotted lines show the performance of SoTA methods without debiasing.}
    \label{fig:oracle_performance}
\end{figure}

To verify our claim, we perform a simulation study to demonstrate the oracle (or upper bound) search performance by removal of spurious correlation. The experiment is carried out by simulating the performance of the multi-label classifier at different levels of prediction accuracy. For example, with an accuracy of 100\%, we expect the potential bias in representation learning to be removed; hence, higher search performance is expected. Figure~\ref{fig:oracle_performance} shows the R1 performance trend with different levels of classification accuracy on the entire test set of Recipe 1M with 50K image-recipe pairs. The performance is obtained by plugging the debiasing module into H-T~\cite{salvador2021revamping}. With 20\% classification accuracy, R1 can reach 44.4\%, which is far better than that of H-T (14.3\%), TFood (25.7\%) and VLPCook (29.7\%). As accuracy becomes higher, the value of R1 also increases and reaches the oracle performance of  91.8\%. The result basically verifies our derivation in Eq.~(\ref{eq:approx_eq2}). By correcting the bias with the probability distribution of ingredients, the debiased image representations can better reflect the ingredient composition. To further verify the performance, we also compare to another oracle run that ranks recipes based on the number of ingredients overlapped between a recipe and the query image. This is a simple technique used in~\cite{chen2016deep} based on text matching. As shown in Figure~\ref{fig:oracle_performance}, when the classification accuracy reaches 100\%,  the R@1 performance is similar to H-T+debiasing. However, R@1 deteriorates quickly when classification accuracy drops. To achieve R@1 of 20\%, the required classification accuracy is 50\%. H-T+debiasing, in contrast, can attain similar performance with only 10\% of classification accuracy. The result also shows the robustness of cross-modal representation learning compared to the classification-based recipe retrieval in~\cite{chen2016deep} using text matching.

\subsection{Comparison to Existing Works}

Table~\ref{tab:comparison} lists the search performances of image-to-recipe retrieval, including our work ``+Debiasing" plugged in H-T, TFood, and VLPCook. On the 1K test set, using the oracle performance as reference, the performances (medR, R5, and R10) have almost saturated for some models. Our proposed debiasing module manages to consistently boost the top-performing models by around 1\% to 4\% of absolute improvement in Recall@1. A similar level of performance boost is also attained on the 10K test set. Particularly, the debiasing module is able to push medR of VLPCook to 1.4, which is the new best result on this dataset. Our results are also better than the approaches based on disentanglement\cite{kim2021learning, sugiyama2021cross} and cooking program generation~\cite{papadopoulos2022learning}, which aim for robust representation learning. Please note that, as the source code of CIP~\cite{huang2023improving} is not publicly available, we are not able to test the effect of debiasing for CIP.

{The F1 score of ingredient classification varies between 21.8\% (for VLPCook) and 37.5\% (for TFood).} It is worth noting that the debiasing module introduces insignificant computation overhead. Typically, only a small number of ingredients, ranging from 1 to 8, will be predicted by the classifier. Retrieving the corresponding ingredient representations from the dictionary to calculate Eq.~(\ref{eq:approx_eq2}) incurs almost no overhead. The full set of result comparisons, including the performances of recipe-to-image retrieval and ingredient classification, can be found in the supplementary.

\begin{table}[]
    \centering
    \caption{Comparison on 1k and 10k test sets for image-to-recipe retrieval. The proposed debiasing successfully boosts the performance of existing cross-modal retrieval methods (H-T, TFood, VLPCook), especially on the 10k set.}
    \begin{adjustbox}{width=\columnwidth}
    \begin{tabular}{@{}l|cccc|cccc@{}}
\toprule

\multirow{2}{*}{} & \multicolumn{4}{c|}{\textbf{1k}}   & \multicolumn{4}{c}{\textbf{10k}} \\ 
                  & medR     & R@1      & R@5    & R@10     & medR     & R@1      & R@5    & R@10     \\ \midrule
                  
RIVAE~\cite{kim2021learning}  & 2.0      & 39.0    & 70.0      &   79.0      &    -   &      -  &     -   &    -    \\





RDE-GAN \cite{sugiyama2021cross}               &    1.0      & 55.1        &     86.7  &    92.4      &    5.0      &    24.0    &    51.6   &    65.4      \\

X-MRS \cite{guerrero2021cross}             &  1.0       &     64.0    &    88.3   &    92.6  &     3.0    &    32.9     &    60.6   &    71.2    \\

Cooking Program ~\cite{papadopoulos2022learning}  &   1.0    &  66.9   &  90.9     &   95.1      &    -   &      -  &     -   &    -    \\

CIP~\cite{huang2023improving}  &    1.0     &   77.1   & 94.2   &   97.2     &  2.0     &   44.9    &   72.8  &  82.0    \\\midrule

H-T~\cite{salvador2021revamping}            &  1.0  &  61.8 &   88.0  & 93.2   &    3.95  &  29.9 & 58.3  &  69.6     \\ 
\textbf{+Debiasing}           &  1.0  & 65.7  &  89.8   & 94.1     &   3.0   &  34.4 &  62.9 &  73.6          \\ \midrule

TFood~\cite{shukor2022transformer}           &  1.0  &  72.4 &   92.5  & 95.4   &   2.0   & 43.9  & 71.7  & 80.8          \\ 
\textbf{+Debiasing}            &  1.0  &  74.5   & 93.2 & 96.1   & 2.0  & 45.6  &  73.0 & 81.6           \\ \midrule

VLPCook~\cite{shukor2022structured}           &  1.0  &  77.4 &   94.8  & 97.1    &   2.0   & 48.8  & 76.2  & 84.5                       \\ 
\textbf{+Debiasing}            &  1.0  & 78.3  &  95.1   & 97.4  &  1.4    & 50.2  & 77.3 & 85.2                   \\ \midrule

Oracle           &  1.0  &  99.0 &  99.8   & 99.9     &  1.0    & 96.2 & 99.2  &  99.5            \\ 
\bottomrule
\end{tabular}
    \end{adjustbox}
    
\label{tab:comparison}
\end{table}

\begin{table}[]
    \centering
    \caption{Scalability test on 20k, 30k, 40k and 50k test set.}
    \begin{adjustbox}{width=\columnwidth}
    \begin{tabular}{@{}l|cc|cc|cc|cc@{}}
\toprule
\multirow{2}{*}{} 
                  & \multicolumn{2}{c|}{\textbf{20k}} & \multicolumn{2}{c|}{\textbf{30k}} & \multicolumn{2}{c|}{\textbf{40k}} & \multicolumn{2}{c}{\textbf{50k}} \\ 
                  & medR     & R@1      & medR     & R@1       & medR     & R@1        & medR     & R@1       \\ \midrule

H-T~\cite{salvador2021revamping}            &  6.3  &  22.2  & 9.0 &  18.4      &  12.0    & 16.0   & 15.0 & 14.3   \\ 
\textbf{+Debiasing}           &  5.0  & 26.2   & 7.0 &  22.0       &   9.0   & 19.3   & 11.0 & 17.4        \\ \midrule

TFood~\cite{shukor2022transformer}           &  3.0  & 35.5  &  4.0 &  30.9      &   5.0   & 27.8 & 6.0  & 25.7       \\ 
\textbf{+Debiasing}            &  3.0  & 37.6   & 3.0 &  32.9      &   4.0   & 29.9   & 5.0  & 26.9          \\ \midrule

VLPCook~\cite{shukor2022structured}          &  2.0  &  40.2  & 3.0 &   35.2    &   4.0   & 32.0  &  4.0 & 29.7                   \\ 
\textbf{+Debiasing}            &  2.0  & 41.7    &  3.0  &  36.9   & 3.0 & 33.7   & 4.0  & 31.1       \\ \midrule

Oracle           &  1.0  & 94.6  & 1.0 & 93.5     &  1.0    & 92.6   & 1.0 & 91.8                    \\ 
\bottomrule
\end{tabular}
    \end{adjustbox}
    
\label{tab:scabulity}
\end{table}

\begin{figure*}
    \centering
    \includegraphics[width=15cm]{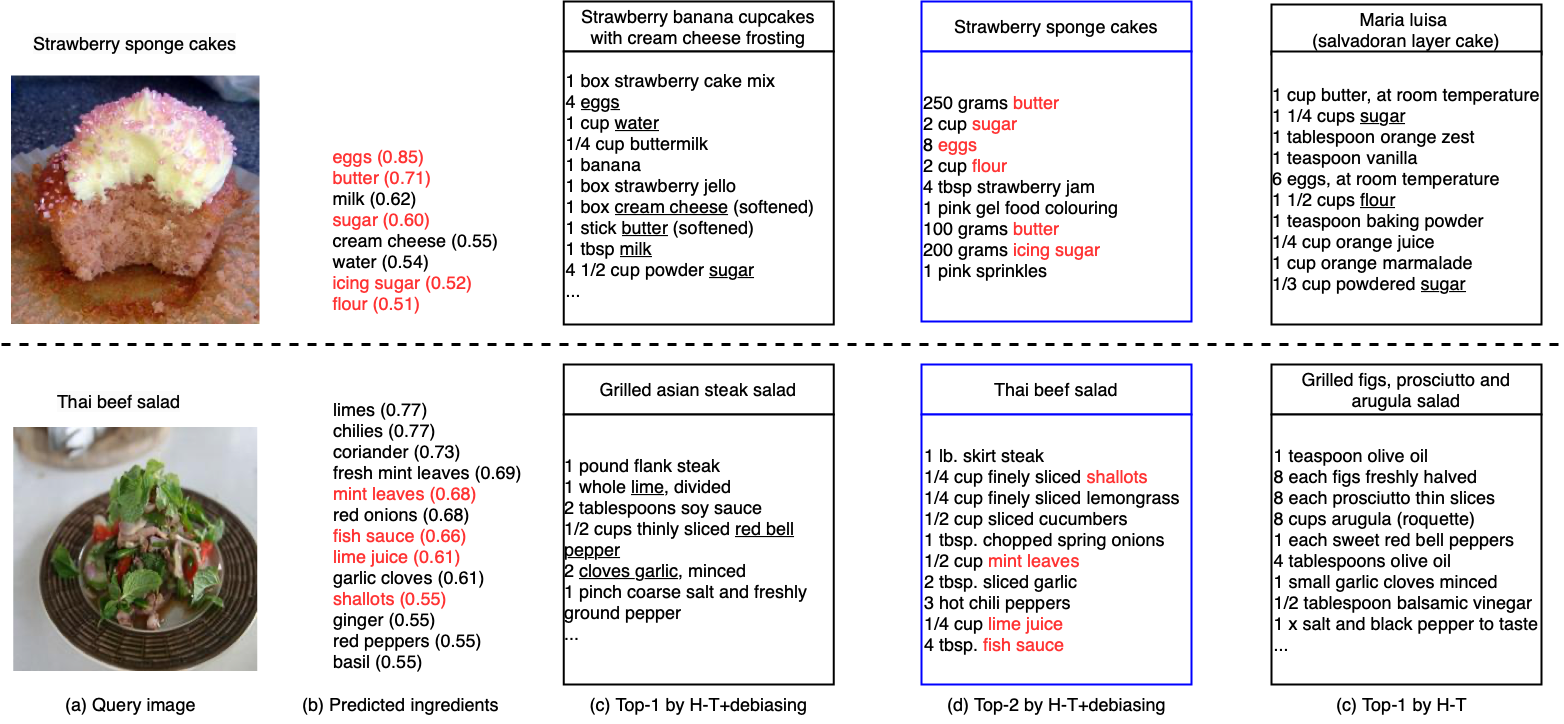}
    \caption{Two examples providing insights on the debiasing mechanism: query image (a), predicted ingredient (b), the retrieved recipes (c)-(e). The ground-truth recipes are boxed in blue. The correct predicted ingredients are marked in red. H-T ranks the ground-truth recipes at the ranks more than 100. Best viewed in color. }
    \label{fig:ret2cls}
\end{figure*}

In Figure~\ref{fig:ret2cls}, two examples are shown to provide insights into how the debiasing module amends the image representation of H-T with the predicted ingredient list. In the first example, around half the ingredients in the ground-truth recipe are recalled correctly. Despite false predictions, the debiased image representation is still able to disambiguate the cake-relevant recipes that are ranked high by H-T. By debiasing, the ground-truth recipe ``Strawberry sponge cakes", which is ranked at $105^{th}$ position by H-T, is boosted to the top-2 rank. Similarly in the second example, the ground-truth recipe ``Thai beef salad" can be boosted from $251^{th}$ to top-2 rank by the debiasing module. The ingredients unique to this recipe, such as limes which are correctly predicted, help to downgrade the rank of other salad recipes, such as ``Grilled figs, prosciutto, and arugula salad" which is retrieved by H-T as the top-1 recipe. Please find more examples in the supplementary document, including failure examples due to incorrect prediction of major ingredients.

\subsection{Robustness Test}\label{sec:exp_robust}



\begin{table}[]
    \centering
    \caption{Median rank comparison for unseen dish categories on the 50k test set.}
    \begin{adjustbox}{width=0.7\columnwidth}
     \begin{tblr}{ccccccc}
\hline
Food type   & Oracle  & H-T &  {H-T\textbf{+Debiasing}} \\ \hline
pizza & 1.0 & 23.0 & 14.0  \\ \hline 
steak & 1.0 & 27.0 & 23.0     \\ \hline 
pancakes & 1.0 & 32.0 & 27.0     \\ \hline 
cheesecake & 1.0 & 29.0 & 17.5    \\ \hline 
cupcake & 1.0 & 22.0 & 17.0      \\ \hline 
lasagna & 1.0 & 18.0 & 13.0    \\ \hline 
fried rice & 1.0 & 15.0 & 8.0      \\ \hline 
tacos & 1.0 & 17.0 & 12.5      \\ \hline 
burger & 1.0 & 23.0 & 15.0   \\ \hline 
waffles & 1.0 & 19.0 & 17.0     \\ \hline 

\end{tblr}
\end{adjustbox}
        \label{tab:zero_shot_food}
\end{table}

\textbf{Scalability}.
Causality-based training theoretically can improve model robustness, as shown in~\cite{wang2021enhancing}. In this section, we present the search results on larger test sets, ranging from 20K pairs to 50K pairs, as shown in Table~\ref{tab:scabulity}. Note that we also show the oracle result as reference, in which the debiasing module is simulated to have a perfect accuracy for ingredient classification. The theoretical upper bound performance clearly shows that the results are robust to the data scale. For example, medR is consistently equal to 1.0 across all data sizes, and the fluctuations in Recall@1 are all less than 3\% absolute difference from data size of 20K to 50K. By plugging in the debiasing module to various models, consistent improvements are also attained. For H-T and TFood, larger margins of improvement are generally noted with the increase in test size. For VLPCook, despite its strong performance even on the large test set, the debiasing module is still able to boost the recall performances. 

\textbf{Zero-Shot Retrieval}.
We also show the robustness of retrieving the unseen food categories (i.e., zero-shot retrieval). In Recipe1M, the recipes are grouped into 1,048 different semantic classes (or food categories). In this experiment, we remove all the recipes grouped under some categories from the training and validation sets for the experiment. For example, the recipes falling under any category with the word ``steak" (e.g., flank steak, grilled steak) or ``pizza" (e.g., pepperoni pizza, cheese pizza) are removed. In total, there are 78 dish categories (involving 14,415 recipes) being removed. We further group these categories based on the removed words and present the search result in Table~\ref{tab:zero_shot_food}. For example, the first row ``pizza" shows the medR performance for all the queries categorized as ``pizza" related in the test set. Note that the oracle performance can still reach 1.0 for all the categories. By debiasing H-T, substantial improvements in medR are attained across all the categories. For example, the medR for ``cheesecake" is improved by more than 10 ranks. This is mainly because H-T always ranks higher for those cakes (e.g., honey cake and desserts) that are seen during training. As the debiasing module is able to predict ingredients such as onion, salt, and black pepper, which are relatively unique for ``cheesecake", the debiased image representation is more robust in ranking the recipes of unseen food categories.

\subsection{Impact of Dictionary}
In theory, all the ingredients should be considered for debiasing. Nevertheless, the oracle performance indeed suggests a near-perfect performance (i.e., Recall@\{1, 5, 10\} = 1.0) even by indexing only the 500 most popular ingredients in the dictionary. This insight is significant for alleviating the need for a large-scale classifier and a large-size dictionary for debiasing. Especially, considering that the multi-label classifier performance is expected to be adversely affected by the increased number of ingredients, this also implies that there is a good trade-off between retrieval and classification performances. Specifically, we can engineer the dictionary size by including a subset of ingredients, in which the classification accuracy will not be negatively impacted while being able to improve retrieval performance.

\begin{table}[]
    \centering
    
\caption{The impact of different dictionaries and their sizes on ingredient classification and recipe retrieval for H-T+debiasing on 10k test size.}
\begin{adjustbox}{width=\columnwidth}
\begin{tabular}{ccccc|cccc}
\hline
\multirow{3}{*}{Size} & \multicolumn{4}{c|}{Our Dictionary}                       & \multicolumn{4}{c}{Inverse Cooking~\cite{salvador2019inverse}}                       \\ \cline{2-9} 
                      & \multicolumn{3}{c}{Classification} & \multirow{2}{*}{Recall@1} & \multicolumn{3}{c}{Classification} & \multirow{2}{*}{Recall@1} \\ \cline{2-4} \cline{6-8}
                      & Precision     & Recall     & F1    &                     & Precision     & Recall     & F1    &                     \\ \hline
100                   & 35.6         & 49.0      & 41.2    & 32.2                     &   24.7            &     73.4       &    37.0   &    35.8                 \\ 
500                   & 30.7         & 38.1      & 34.0    & 34.4                    &     25.4          &      69.6      &   37.2    &    36.1                 \\ 
1000                  & 29.7         & 35.2      & 32.2    & 34.0                     &     24.5          &     67.1       &   35.9    &   36.1                  \\ \hline
\end{tabular}
\end{adjustbox}
        \label{tab:dic_size}
\end{table}

Table~\ref{tab:dic_size} (left) shows the impact of dictionary size on the performances. In the experiment, we sort the ingredients according to their appearance frequencies on Recipe1M and then keep the top most popular ingredients in the dictionary. We do not exclude ingredients, such as salt and sugar, which may be invisible in a dish. As shown in Table~\ref{tab:dic_size}, having a small-size dictionary generally results in lower retrieval performance despite higher classification performance. Increasing the size to include the 1,000 most popular ingredients, nevertheless, hurts both classification and retrieval performances. A dictionary with 500 ingredients appears to be a good trade-off in our experiment. 

We also train our framework using the dictionary shared by Inverse Cooking~\cite{salvador2019inverse}. The ingredients are predicted using the auto-regressive model~\cite{salvador2019inverse}. Unlike our approach where each ingredient is considered individually, Inverse Cooking merges similar ingredients into one category. For example, in Recipe1M, there are 494 kinds of cheese (e.g., american cheese, cheddar cheese, cream cheese), which are merged into one category. However, in our ingredient dictionary, we do not merge different types of cheeses, and instead, include only the frequent cheese types as separate entries in the dictionary. Table~\ref{tab:dic_size} contrasts the performances between using Inverse Cooking and our dictionaries. As Inverse Cooking covers a broader range of ingredients due to the merging of ingredient types, the retrieval performance is also improved by $1.7\%$ compared to using our dictionary. The result provides insights that the debiasing module can further boost the retrieval performance with a properly curated dictionary considering both the scope and frequency of ingredients. The full set of ablation studies is presented in supplementary.

\begin{figure}
    \centering
    \includegraphics[width=7cm]{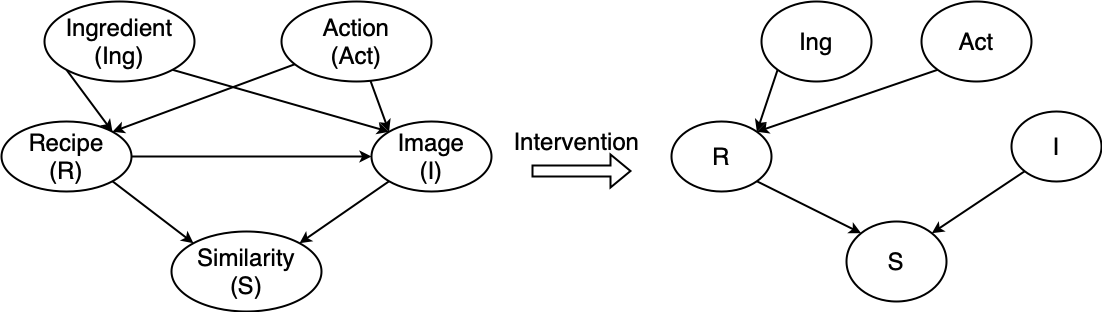}
    \caption{Causal graph with both ingredients and cooking actions as the confounder sources.}
    \label{fig:causal_graph_ing_act}

\end{figure}

\section{Conclusion}

We have presented a causal formulation to remove bias in representation learning on cross-modal paired data. Empirically through the simulation result, we also demonstrate that a near-perfect retrieval performance is attainable by modeling ingredients as the confounder. This implication is important as the results clearly indicate that causality-based representation learning is a promising direction in pursuing high-recall retrieval performance. Our proposed neural module for the causal formula has also consistently boosted the search performance of three SoTA models, achieving the highest reported MedR and recall performances on Recipe1M. Compared to the oracle performance, the proposed module is bottlenecked by the performance of the multi-label classifiers in predicting the distribution of culinary elements. Our future work includes the proposal of more sophisticated classifiers for debiasing.

\appendix

\section*{Appendix}

In this supplementary document, we provide a more comprehensive list of performance comparisons for both image-to-recipe retrieval and recipe-to-image retrieval (Section~\ref{sec:sup_sota_comparison}). In addition, more details of the scalability test, including the ingredient classification performances, are also provided (Section~\ref{sec:robustness_s}). Ablation studies provide further insights into the choice of ingredients for dictionary construction (Section~\ref{sec:dic_cons}) and parameter setting in model training (Section~\ref{sec:sup_lambda}). Finally, we provide more examples for qualitative analysis (Section~\ref{sec:sup_qualitative}).

\setlength{\tabcolsep}{7pt}
\begin{table*}[th]

\centering
\caption{Comparison on 1k and 10k test sets. medR ($\downarrow$), Recall@k ($\uparrow$) are reported. The proposed debiasing successfully boosts the performance of existing cross-modal retrieval methods (H-T, TFood, VLPCook), especially on the 10k set.}
\resizebox{\textwidth}{!}{
\begin{tabular}{@{}l|cccc|cccc|cccc|cccc@{}}
\toprule
\multirow{3}{*}{} & \multicolumn{8}{c|}{\textbf{1k}}      & \multicolumn{8}{c}{\textbf{10k}} \\ \cmidrule(l){2-17} 
                  & \multicolumn{4}{c|}{\textbf{image-to-recipe}} & \multicolumn{4}{c|}{\textbf{recipe-to-image}} & \multicolumn{4}{c|}{\textbf{image-to-recipe}} & \multicolumn{4}{c}{\textbf{recipe-to-image}} \\ \cmidrule(l){2-17} 
                  & medR     & R@1      & R@5    & R@10    & medR     & R@1     & R@5     & R@10    & medR     & R@1      & R@5    & R@10    & medR     & R@1     & R@5     & R@10    \\ \midrule

RIVAE~\cite{kim2021learning}  & 2.0      & 39.0    & 70.0      &   79.0     &   -       & -       &     -   &       - & -   & -     &  -     &   -     &    -   &      -  &     -   &    -    \\

R2GAN \cite{zhu2019r2gan}                  &     2.0     &     39.1    &   71.0    &  81.7      &  2.0        &    40.6    &    72.6    &       83.3 &  13.9        &   13.5      &   33.5    &     44.9   &     12.6     &   14.2     &   35.0     &   46.8     \\
MCEN \cite{fu2020mcen}               &   2.0       &   48.2      &    75.8   &      83.6  &    1.9      &   48.4     &  76.1      &    83.7    &  7.2        &      20.3   &    43.3   &   54.4     &     6.6     &   21.4     &  44.3      &    55.2    \\
ACME \cite{wang2019learning}              &    1.0      &   51.8      &   80.2    &    87.5    &   1.0       &    52.8    &   80.2     &       87.6 &     6.7     &   22.9     &  46.8     &    57.9    &      6.0    &  24.4      &    47.9    &      59.0  \\
SN \cite{zan2020sentence}               &    1.0      & 52.7        &     81.7  &    88.9    &    1.0      &  54.1      &  81.8      &   88.9     &    7.0      &    22.1    &    45.9   &    56.9    &    7.0      &     23.4   &  47.3      &  57.9      \\
IMHF \cite{li2021cross}               &    1.0      & 59.4        &     81.0  &    87.4    &    1.0      &  61.2      &  81.0      &   87.2     &    3.5      &    36.0    &    56.1   &    64.4    &    3.0      &     38.2   &  57.7      &  65.8      \\
SCAN \cite{wang2021cross}               &    1.0      & 54.0        &     81.7  &    88.8    &    1.0      &  54.9      &  81.9      &   89.0     &    5.9      &    23.7    &    49.3   &    60.6    &    5.1      &     25.3   &  50.6      &  61.6      \\
HF-ICMA \cite{li2021hybrid}               &    1.0      & 55.1        &     86.7  &    92.4    &    1.0      &  56.8      &  87.5      &   93.0     &    5.0      &    24.0    &    51.6   &    65.4    &    4.2      &     25.6   &  54.8      &  67.3      \\

MSJE \cite{xiez2021learning}               &    1.0      & 56.5        &     84.7  &    90.9    &    1.0      &  56.2      &  84.9      &   91.1     &    5.0      &    25.6    &    52.1   &    63.8    &    5.0      &     26.2   &  52.5      &  64.1      \\

SEJE \cite{xie2021learning}               &    1.0      & 58.1        &     85.8  &    92.2    &    1.0      &  58.5      &  86.2      &   92.3     &    4.2      &    26.9    &    54.0   &    65.6    &    4.0      &     27.2   &  54.4      &  66.1      \\

M-SIA \cite{li2021multi} &  1.0       &     59.3    &    86.3   &    92.6    &     1.0     &     59.8   &    86.7    &      92.8  &     4.0    &    29.2     &    55.0   &    66.2    &    4.0      &     30.3   &     55.6   &  66.5     \\

RDE-GAN \cite{sugiyama2021cross}               &    1.0      & 55.1        &     86.7  &    92.4    &    1.0      &  56.8      &  87.5      &   93.0     &    5.0      &    24.0    &    51.6   &    65.4    &    4.2      &     25.6   &  54.8      &  67.3      \\

X-MRS \cite{guerrero2021cross}             &  1.0       &     64.0    &    88.3   &    92.6    &     1.0     &     63.9   &    87.6    &      92.6  &     3.0    &    32.9     &    60.6   &    71.2    &    3.0      &     33.0   &     60.4   &  70.7       \\

Cooking Program ~\cite{papadopoulos2022learning}  &   1.0    &  66.8   &  89.8     &   94.6      &    -   &      -  &     -   &    - &    -   &      -  &     -   &    - &    -   &      -  &     -   &    -    \\

CIP~\cite{huang2023improving}  &    1.0     &   77.1   & 94.2   &   97.2     &  1.0    &   77.3     &    94.4   &     97.0   &  2.0     &   44.9    &   72.8  &  82.0    &  2.0   &   45.2   &    73.0    &   81.8   \\\midrule

H-T~\cite{salvador2021revamping}            &  1.0  &  61.8 &   88.0  & 93.2 & 1.0 & 62.1   & 88.3  &  93.5     &    3.95  &  29.9 & 58.3  &  69.6 & 3.6 &   30.4    &  58.6     &      69.7     \\ 
\textbf{+Debiasing}           &  1.0  & 65.7  &  89.8   & 94.1 & 1.0 &  66.0  & 89.9  &  94.2     &   3.0   &  34.4 &  62.9 &  73.6 & 3.0 &  34.7     &   63.2    &   73.7              \\ \midrule

TFood~\cite{shukor2022transformer}           &  1.0  &  72.4 &   92.5  & 95.4 & 1.0 &  72.5  & 92.1  &   95.3    &   2.0   & 43.9  & 71.7  & 80.8  & 2.0 &    43.7   &    71.6   &      80.6           \\ 
\textbf{+Debiasing}            &  1.0  &  74.5   & 93.2 & 96.1 &  1.0  &  73.7 &  93.1     &  96.0    & 2.0  & 45.6  &  73.0 & 81.6 &   2.0    &   44.9   &   72.7  &  81.5              \\ \midrule

VLPCook~\cite{shukor2022structured}           &  1.0  &  77.4 &   94.8  & 97.1 & 1.0 &  78.0  & 94.9  &   97.1    &   2.0   & 48.8  & 76.2  & 84.5  & 1.6 & 49.9 &  76.9     &  85.0                      \\ 
\textbf{+Debiasing}            &  1.0  & 78.3  &  95.1   & 97.4 & 1.0 &  78.6  &  95.2 &  97.4    &  1.4    & 50.2  & 77.3 & 85.2  & 1.0 & 51.0 &  77.9   &  85.6                     \\ \midrule

Oracle           &  1.0  &  99.0 &  99.8   & 99.9 & 1.0 & 98.9   &  99.8 &  99.9     &  1.0    & 96.2 & 99.2  &  99.5 & 1.0 & 96.1 & 99.1      &     99.5                 \\ 
\bottomrule
\end{tabular}
}

\label{tab:comparison_full}
\end{table*}

\section{Recipe-to-Image Retrieval Results} \label{sec:sup_sota_comparison}
The complete comparison to the existing works~\cite{kim2021learning, xiez2021learning, xie2021learning, li2021multi, sugiyama2021cross, guerrero2021cross, papadopoulos2022learning, huang2023improving, salvador2021revamping, shukor2022transformer, shukor2022structured} on the 1K and 10 datasets is shown in Table~\ref{tab:comparison_full}. For both image-to-recipe (I2R) retrieval and recipe-to-image (R2I) retrieval, the proposed debiasing module consistently boosts the performances of H-T~\cite{salvador2021revamping}, TFood~\cite{shukor2022transformer} and VLPCook~\cite{shukor2022structured}. New state-of-the-art performances are achieved by VLPCook + Debiasing. On 10K search dataset, medR= 1.4 for I2R and medR=1.0 for R2I. The result is also much better than RDE-GAN~\cite{sugiyama2021cross}, which performs disentanglement to separate non-recipe shape features from representation learning. For example, on 10K dataset, RDE-GAN attains R1=24.0\% versus R1=50.2\% by VLPCook+Debiasing. This result is also better than the recently published CIP~\cite{huang2023improving}, which explores CLIP and prompts learning for retrieval.

\begin{table}[]
    \centering
    \caption{Impact of dictionary size and visibility of ingredients (on size of 500 ingredients). The table shows the ingredient classification and retrieval performances for H-T+Debiasing on 10k test size. Note that the columns marked with (visible only) show the results of using a dictionary that includes only ingredients that will likely be visible in a final cooked dish.}
    \begin{adjustbox}{width=\columnwidth}

\begin{tabular}{ccccc}
\hline
\multirow{2}{*}{Size} & \multicolumn{3}{c}{Classification} & \multirow{2}{*}{Recall@1} \\ \cline{2-4}
                      & Precision    & Recall    & F1      &                           \\ \hline
100                   & 35.6         & 49.0      & 41.2    & 32.2                      \\ \hline
250 (Visible only) & 30.8 & 37.5 & 33.8 &  34.0 \\ \hline 

500                   & 30.7         & 38.1      & 34.0    & 34.4                      \\ \hline
500 (Visible only) & 29.1 & 33.9 & 31.3 & 34.3  \\ \hline 

1000                  & 29.7         & 35.2      & 32.2    & 34.0                      \\ \hline
\end{tabular}
   
    \end{adjustbox}{}
     
        \label{tab:dic_size_x}
\end{table}

\begin{table*}[th]
\centering
\caption{Scalability test on 20k, 30k, 40k and 50k test set for the image-to-recipe retrieval task.}
\resizebox{\textwidth}{!}{
\begin{tabular}{@{}l|cccc|cccc|cccc|cccc@{}}
\toprule
\multirow{2}{*}{} 
                  & \multicolumn{4}{c|}{\textbf{20k}} & \multicolumn{4}{c|}{\textbf{30k}} & \multicolumn{4}{c|}{\textbf{40k}} & \multicolumn{4}{c}{\textbf{50k}} \\ \cmidrule(l){2-17} 
                  & medR     & R@1      & R@5    & R@10    & medR     & R@1     & R@5     & R@10    & medR     & R@1      & R@5    & R@10    & medR     & R@1     & R@5     & R@10    \\ \midrule

H-T~\cite{salvador2021revamping}            &  6.3  &  22.2 &  47.0   & 58.8 & 9.0 &  18.4  & 41.1  &  52.5     &  12.0    & 16.0 &  36.9 & 47.9  & 15.0 & 14.3 &  33.8     &  44.4  \\ 
\textbf{+Debiasing}          &  5.0  & 26.2  &  52.4   & 63.7 & 7.0 &  22.0  &  46.2 &  57.7     &   9.0   & 19.3 &  41.9 & 53.2  & 11.0 & 17.4 &   38.7    &    49.6       \\ \midrule

TFood~\cite{shukor2022transformer}           &  3.0  & 35.5  &  62.0   & 72.5 &  4.0 &  30.9  &  56.0 & 66.7      &   5.0   & 27.8 &  52.2 &  62.8 & 6.0  & 25.7 & 49.1  &  59.7       \\ 
\textbf{+Debiasing}            &  3.0  & 37.6  &  64.3   & 73.9 & 3.0 &  32.9  & 58.6  &  69.0     &   4.0   & 29.9 &  54.5 & 65.1  & 5.0  & 26.9 & 51.2  & 61.5         \\ \midrule

VLPCook~\cite{shukor2022structured}          &  2.0  &  40.2 &  67.4   & 77.2 & 3.0 &   35.2 &  61.6 &  72.2     &   4.0   & 32.0 &  57.5 &  68.4 &  4.0 & 29.7 & 54.5  & 65.3                  \\ 
\textbf{+Debiasing}          &  2.0  & 41.7  &  69.1   & 78.5  &  3.0  &  36.9 &   63.5    &   73.5   & 3.0 & 33.7  & 59.7 & 69.9 & 4.0  & 31.1 & 56.4  & 66.7       \\ \midrule

Oracle           &  1.0  & 94.6  &   98.8  & 99.2 & 1.0 & 93.5   & 98.4  &    99.0   &  1.0    & 92.6 & 98.0  & 98.8  & 1.0 & 91.8 &    97.7   &  98.7                    \\ 
\bottomrule
\end{tabular}
}

\label{tab:scabulity_full}
\end{table*}

\section{Scalability Test} \label{sec:robustness_s}

\begin{figure}
    \centering
    \includegraphics[width=8.5cm]{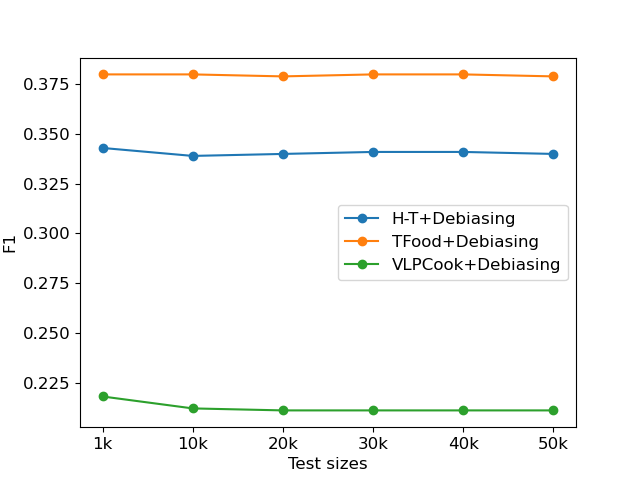}
    \caption{Performance of ingredient prediction remains stable over different data sizes.}
    \label{fig:cls_perf}
\end{figure}

Table~\ref{tab:scabulity_full} lists the full results of image-to-recipe retrieval across different data sizes (20K to 50K). Note that the oracle results are robust over different data scales and performance metrics. The proposed debiasing module consistently improves H-T~\cite{salvador2021revamping}, TFood~\cite{shukor2022transformer} and VLPCook~\cite{shukor2022structured}. There are still performance gaps between the actual and oracle performances. The gaps are mainly due to the prediction accuracy of the ingredient classifier, which is shown in Figure~\ref{fig:cls_perf}. For H-T and TFood, the F1 performance is in the range of 30\% to 40\% across different data sizes. The F1 performance of VLPCook is relatively lower. Although VLPCook explores CLIP to augment the context information for representation learning, these additional contexts do not boost the F1 of ingredient prediction. Despite the lower performance in ingredient prediction, the debiasing module is still able to introduce search improvement for VLPCook.

\section{Ingredient Dictionary} \label{sec:dic_cons}

The dictionary is composed of popular ingredients in Recipe1M, including those ingredients which could become ``invisible" (e.g., salt, butter) during cooking. Intuitively, ingredients, that are not visible in a food image, are not likely to be predicted and hence may be redundant if being included in a dictionary. Table~\ref{tab:dic_size_x} shows the empirical insights of this intuition. We first remove 250 invisible ingredients from the default dictionary of size 500. The retrieval performance is slightly impacted. By adding another 250 visible ingredients (based on their frequencies) to this dictionary, the retrieval performance is slightly improved, but not better than the default dictionary with both visible and invisible ingredients. In general, we note that invisible ingredients still play a supplementary role in debiasing the image representation. For example, the lime juice (the second example in Figure 4 of the main paper.) helps retrieve the right salad recipe. We believe that the classifier has learned to infer ingredients that are supposedly hard to be or not visible in an image from the co-occurrence relationship of ingredients in cooking~\cite{chen2020zero}.

\section{Influence of $\lambda_{cls}$} \label{sec:sup_lambda}

\begin{figure}
    \centering
    \includegraphics[width=8.5cm]{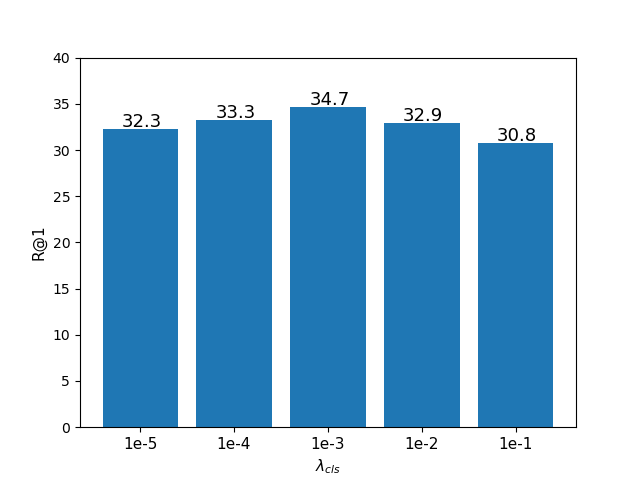}
    \caption{The performance of image-to-recipe retrieval when we vary the value of $\lambda_{cls}$. We report R@1 on a test size of 10k.}
    \label{fig:ablatoion_cls_weight}
\end{figure}

The value of $\lambda_{cls}$ has a moderate influence on the training of our retrieval model. In Figure~\ref{fig:ablatoion_cls_weight}, an inappropriate $\lambda_{cls}$ setting hurts the model performance by around 1\% in R@1. This is because, with a large $\lambda_{cls}$, the classification may bring more perturbation into joint embedding learning. On the other hand, a small $\lambda_{cls}$ setting will weaken the effects of the classification task in representation learning. Therefore we fix $\lambda_{cls}=1e^{-3}$ in our experiments.

\section{Qualitative Analysis} \label{sec:sup_qualitative}

\begin{figure}
    \centering
    \includegraphics[width=8cm]{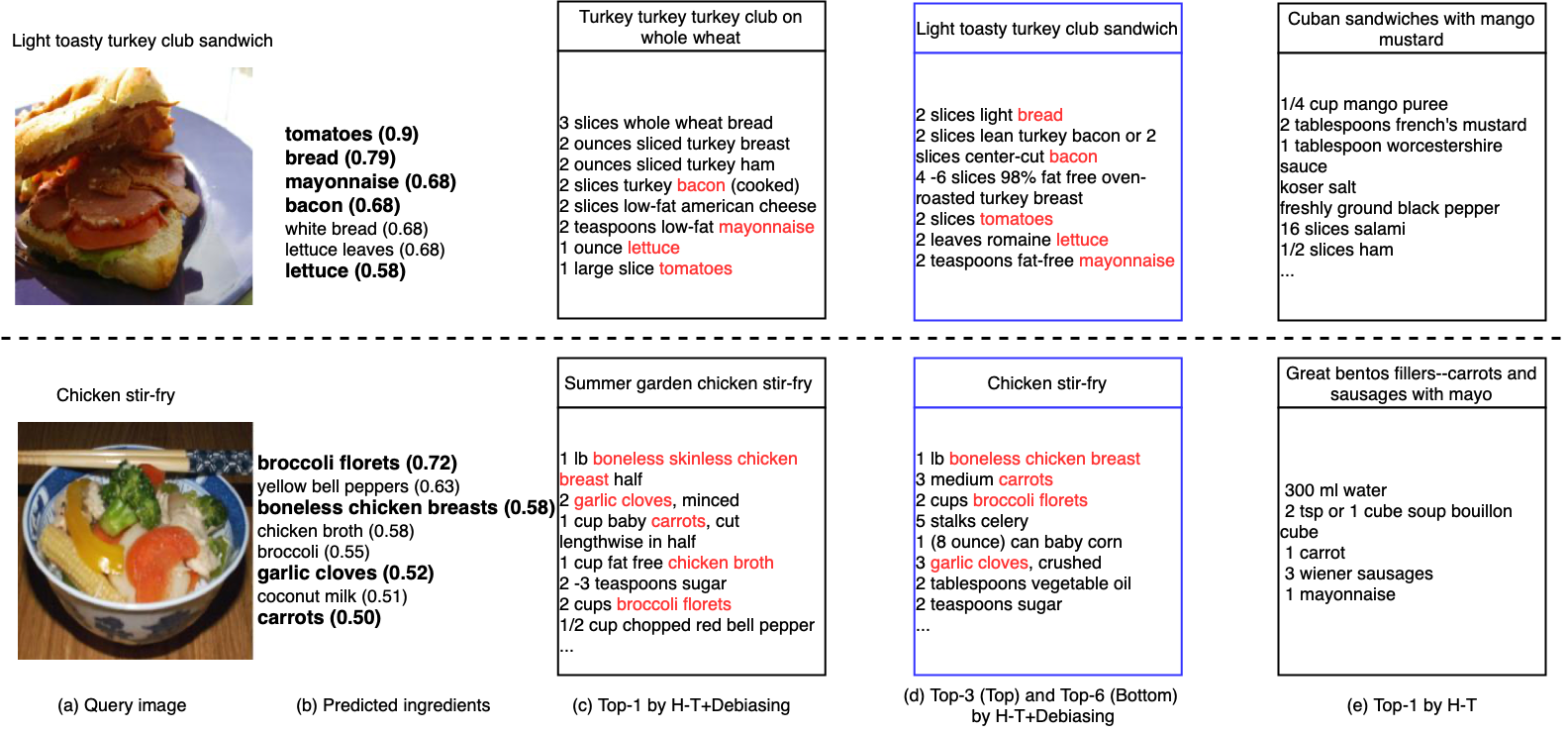}
    \caption{Examples providing insights on the debiasing mechanism: query image (a), predicted ingredient (b), the recipes retrieved by H-T+Debiasing (c), and H-T (d). The ingredients in red are the predicted ingredients as well as ingredients that appeared in recipes retrieved by the method using debiasing.}
    \label{fig:ret2cls_ht}
\end{figure}

\begin{figure}
    \centering
    \includegraphics[width=8cm]{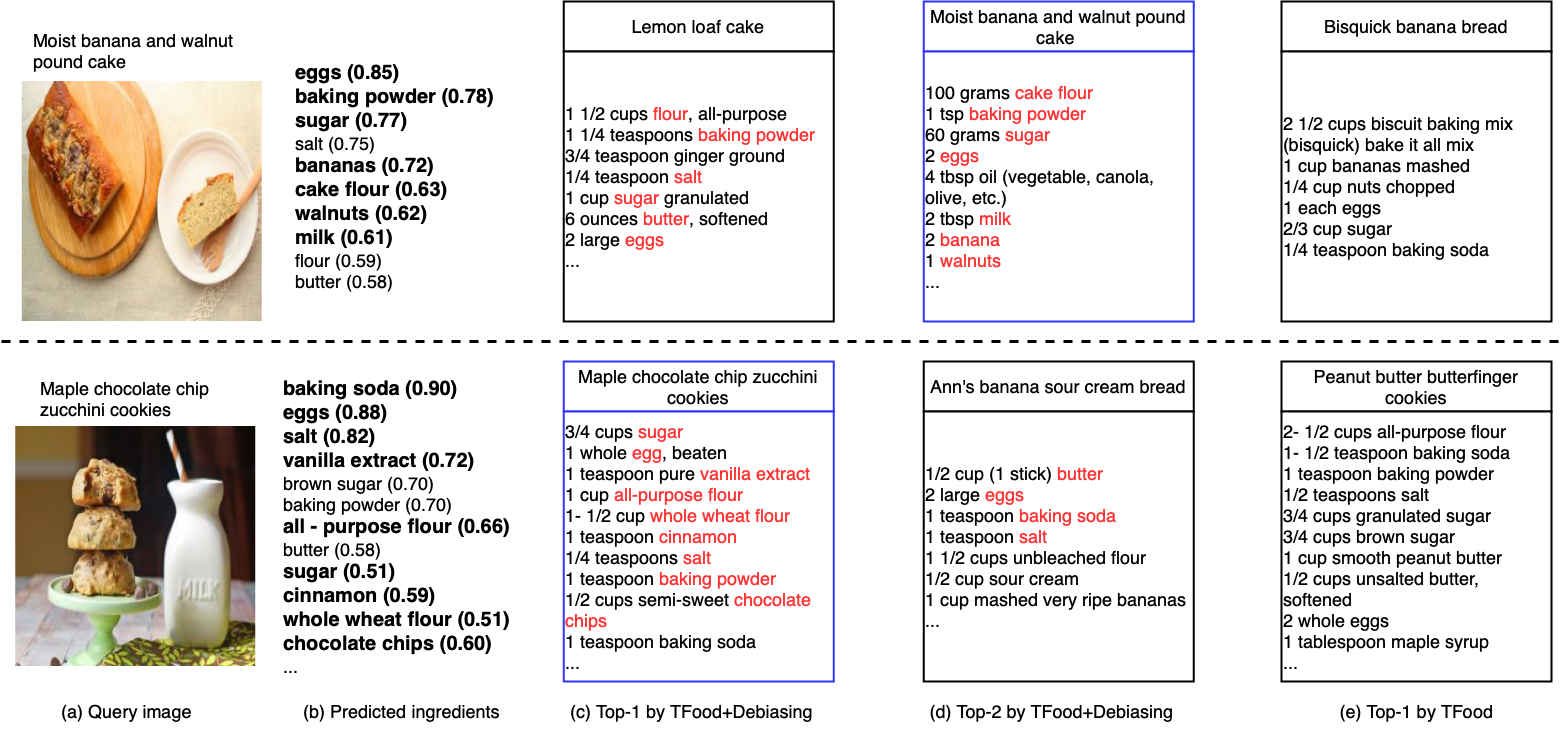}
    \caption{Examples providing insights on the debiasing mechanism: query image (a), predicted ingredient (b), the recipes retrieved by TFood+Debiasing (c), and TFood (d).}
    \label{fig:ret2cls_tfood}
\end{figure}

\begin{figure}
    \centering
    \includegraphics[width=8cm]{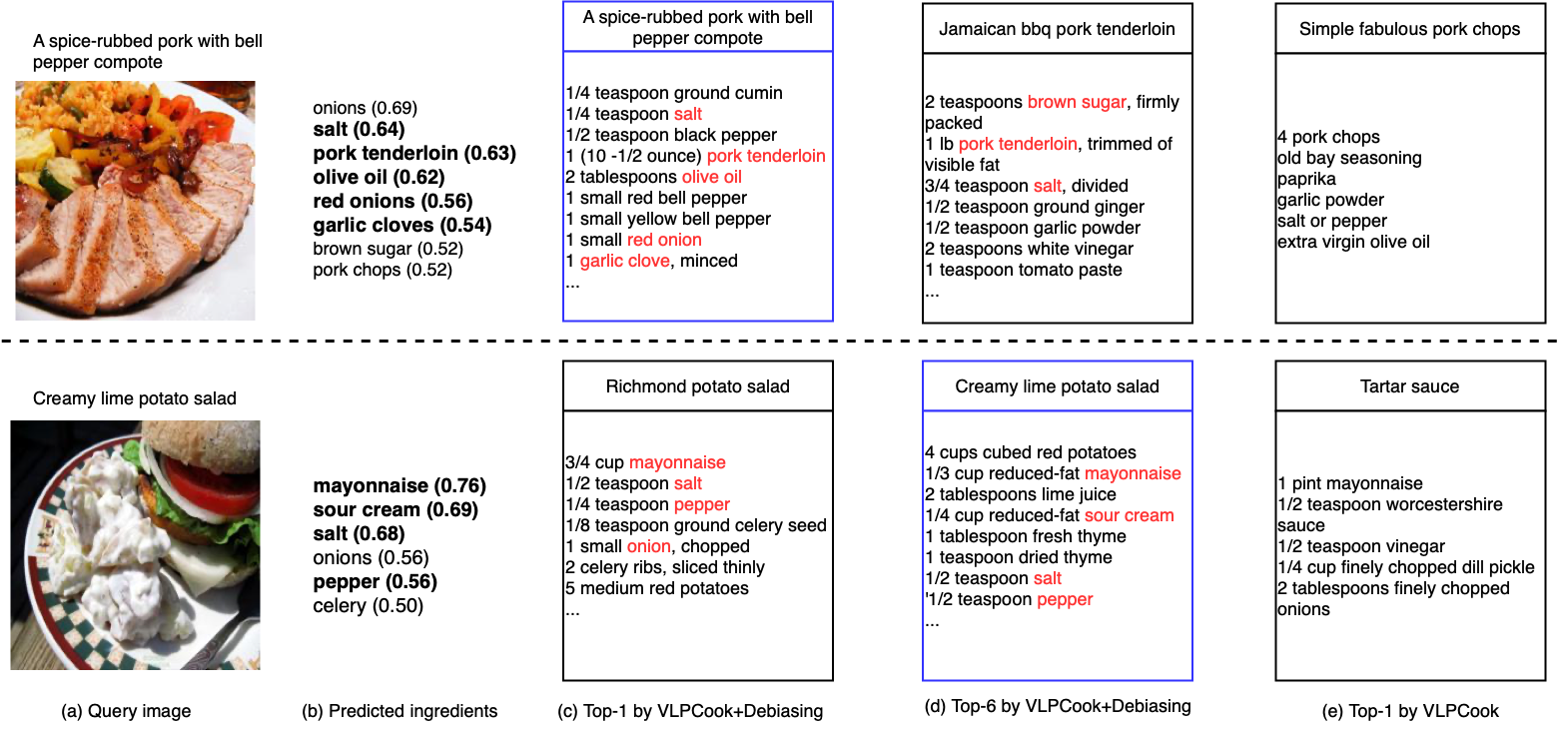}
    \caption{Examples providing insights on the debiasing mechanism: query image (a), predicted ingredient (b), the recipes retrieved by VLPCook+Debiasing (c), and VLPCook (d).}
    \label{fig:ret2cls_vlpcook}
\end{figure}

Figure~\ref{fig:ret2cls_ht},~\ref{fig:ret2cls_tfood},~\ref{fig:ret2cls_vlpcook} show examples explaining how the image representations are ``corrected" with the subtle ingredients predicted by the diabiasing module. In the first example shown in Figure~\ref{fig:ret2cls_ht}, due to the unique dish presentation of burger, the ingredients such as tomato slice and lettuce are occluded. By having the debiasing module to predict these ingredients correctly and add them back to the image representation, the rank of the ground-truth recipe is boosted from $131^{th}$ position (by H-T) to $3^{rd}$ position (by H-T+debiasing). Similarly, for the second example, the boneless chicken breasts are occluded by other ingredients and the garlic cloves are hardly observed in the query image. By having the debiasing to correctly predict them, the ground-truth recipe is boosted from $268^{th}$ rank (by H-T) to $6^{th}$ rank. 

Figure~\ref{fig:ret2cls_tfood} illustrates how TFood+debiasing is able to disambiguate recipes having similar major ingredients. In the first example, the debiasing module predicts the ingredients, such as walnut and cake flour, which are essential to disambiguate the recipe ``moist banana and walnut pound cake" from other cake-like recipes with banana as the main ingredient. The ground-truth recipe is ranked at $2^{nd}$ position versus $561^{th}$ position by T-Food. Similarly in the second example, with the correct prediction of minor ingredients, such as chocolate chips and cinnamon, T-Food+Debiasing is able to disambiguate cooking recipes and retrieves the ground-truth ``maple chocolate chip zucchini cookies" successfully. Figure~\ref{fig:ret2cls_vlpcook} shows the result of boosting VLPCook search performance by the debiasing module. Basically, the ability to debiase image representation, through augmenting a representation with the minor ingredients predicted from a query dish, is helpful in disambiguating recipes, especially for those dishes that share a similar visible appearance or main ingredients.


\begin{figure*}
    \centering
    \includegraphics[width=16cm]{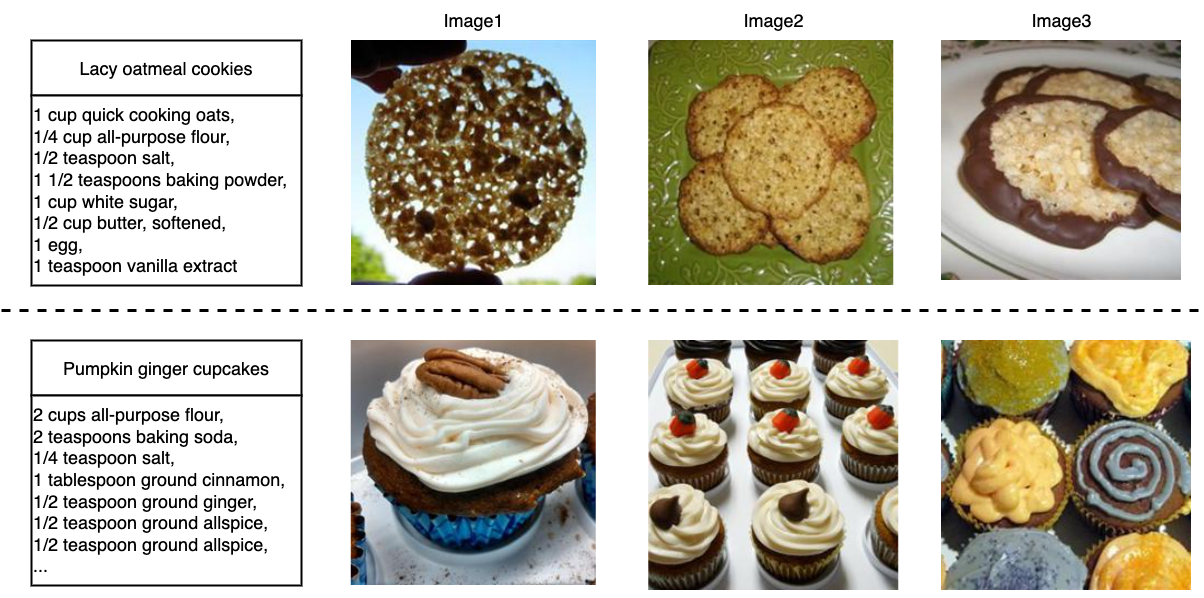}
    \caption{The image examples paired with a recipe. These images are uploaded by users who cook the dishes based on the same recipes. As shown, these images are visually different even though they are prepared with the same recipes.}
    \label{fig:ret2cls_3in1}
\end{figure*}

\begin{figure*}
    \centering
    \includegraphics[width=14.3cm]{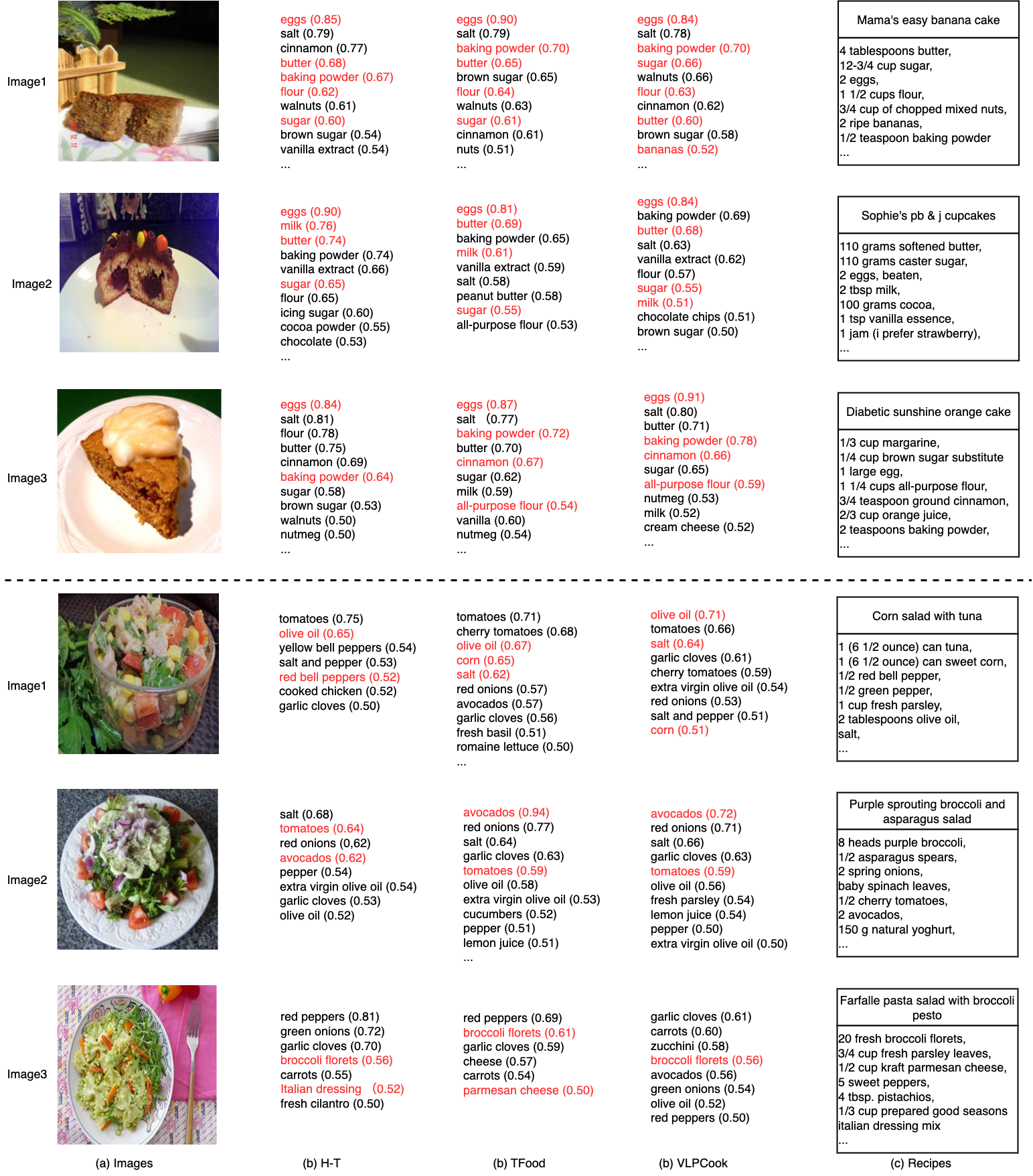}
    \caption{Examples of recipes with similar food images (due to food texture and color). Top: three different recipes of cake. Bottom: salad recipes. The predicted ingredients enable more precise retrieval of the recipes (see results in Table~\ref{tab:ret2cls_sim_3in1}). The ingredients predicted by different models are shown in the middle three columns.}
    \label{fig:ret2cls_sim_3in1}
\end{figure*}

Next, we further highlight the challenge of Recipe1M dataset: high intra-class variation (i.e., visually different food images associated with a recipe, see Figure~\ref{fig:ret2cls_3in1}); low inter-class variation (i.e., visual similar dishes for different recipes, see Figure~\ref{fig:ret2cls_sim_3in1}). Figure~\ref{fig:ret2cls_3in1} shows three recipes, each is paired with three food images. These images are uploaded by different users, and hence, differ in camera viewing angle, dish presentation (e.g., number of items on a plate), and even the color-texture of a final dish. Due to different viewing angles, the shape of a food item may be visible in one image (e.g., the second image of ``pumpkin ginger cupcakes") but not in another image (e.g., the third image). This basically shows the challenge of recipe retrieval, as in Table~\ref{tab:ret2cls_3in1} which lists the ranks of recipes retrieved by different models. For the cookie recipe, note that the retrieval performances for three image queries are largely different. By having a debiasing model, the retrieval performances for these query images are improved across all the models. Especially, our ``HT+Debiasing" can boost the rank of the ground-truth recipe ``pumpkin ginger cupcakes'' to the top-1 position for the first and second query images.

\begin{table*}[]
\centering
\resizebox{\textwidth}{!}{
\begin{tabular}{cccccccc}
\hline
\multicolumn{2}{c}{\multirow{2}{*}{}} & \multicolumn{2}{c}{H-T}      & \multicolumn{2}{c}{TFood}    & \multicolumn{2}{c}{VLPCook}  \\ \cline{3-8} 
\multicolumn{2}{c}{}                  & w/o debiasing & w/ debiasing & w/o debiasing & w/ debiasing & w/o debiasing & w/ debiasing \\ \hline
\multirow{3}{*}{Lacy oatmeal cookies}   & Image1   & 1719          & 724          & 878           & 805          & 71            & 18           \\ \cline{2-8} 
                           & Image2   & 68            & 41           & 15            & 14           & 10            & 6            \\ \cline{2-8} 
                           & Image3   & 1166          & 231          & 468           & 276          & 365           & 127          \\ \hline
\multirow{3}{*}{Pumpkin ginger cupcakes}   & Image1   & 100           & 1            & 44            & 16           & 33            & 3            \\ \cline{2-8} 
                           & Image2   & 153           & 1            & 70            & 19           & 46            & 13           \\ \cline{2-8} 
                           & Image3   & 357           & 298          & 441           & 336          & 10            & 9            \\ \hline
\end{tabular}
}
\caption{The positions of recipes in their respective rank lists retrieved by three different models with (w/) and without (w/o) debiasing modules. The two groups of cookies and cupcakes images are shown in Figure~\ref{fig:ret2cls_3in1}.}
\label{tab:ret2cls_3in1}
\end{table*}

\begin{table*}[]
\centering
\resizebox{\textwidth}{!}{
\begin{tabular}{clcccccc}
\hline
\multicolumn{2}{c}{\multirow{2}{*}{}}                                   & \multicolumn{2}{c}{H-T}      & \multicolumn{2}{c}{TFood}    & \multicolumn{2}{c}{VLPCook}  \\ \cline{3-8} 
\multicolumn{2}{c}{}                                                    & w/o debiasing & w/ debiasing & w/o debiasing & w/ debiasing & w/o debiasing & w/ debiasing \\ \hline
\multirow{3}{*}{Cakes}  & mama's easy banana cake                       & 1197          & 682          & 67            & 47           & 5             & 2            \\ \cline{2-8} 
                        & sophie's pb \& j cupcakes                     & 26            & 8            & 116           & 3            & 2             & 2            \\ \cline{2-8} 
                        & diabetic sunshine orange cake                 & 104           & 85           & 149           & 76           & 125           & 34           \\ \hline
\multirow{3}{*}{Salads} & corn salad with tuna                          & 106           & 4            & 6             & 4            & 2             & 1            \\ \cline{2-8} 
                        & purple sprouting broccoli and asparagus salad & 257           & 120          & 68            & 11           & 1             & 1            \\ \cline{2-8} 
                        & farfalle pasta salad with broccoli pesto      & 332           & 7            & 1             & 1            & 2             & 1            \\ \hline
\end{tabular}
}
\caption{The positions of recipes in their respective rank lists retrieved by three different models with (w/) and without (w/o) debiasing modules. The two groups of cake and salad images are shown in Figure~\ref{fig:ret2cls_sim_3in1}.}
\label{tab:ret2cls_sim_3in1}
\end{table*}

Figure~\ref{fig:ret2cls_sim_3in1} shows two sets of recipes (cake and salad) that exhibit relatively low inter-class visual variation. The food images are somewhat similar in terms of shape (cake) and texture (salad) despite the different composition of ingredients as listed in their recipes. For example, in the first set, bananas and orange juice are only found in the first and third recipes respectively. To show the advantage of the debiasing module, Table~\ref{tab:ret2cls_sim_3in1} lists the ranks of these recipes by using the images in Figure~\ref{fig:ret2cls_sim_3in1} as the queries. As shown, our ``+Debiasing" consistently boosts the ranks of these recipes across three models, especially if the ingredients unique to a recipe are correctly predicted. For example, the ground-truth recipe of ``diabetic sunshine orange cake" is boosted to rank $34^{th}$ when debiasing module manages to predict cinnamon for VLPCook+Debiasing. Similarly, the groundtruth recipe of ``farfalle pasta salad with broccoli pesto" is also boosted to $7^{th}$ position (H-T) when the debiasing module predicts the ingredient Italian dressing.

{\small
\bibliographystyle{ieee_fullname}
\bibliography{WAFT}
}

\end{document}